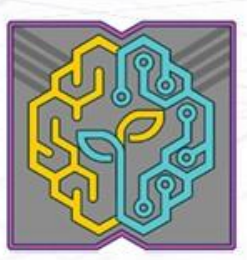
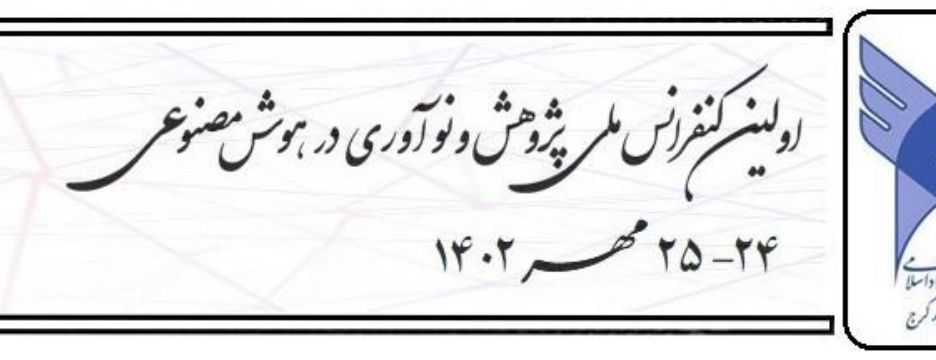

# The Entropy Triangle Method (ETM): A novel framework for the prevention of cardiac arrhythmia with a review of more than 10,000 patients

Arman Daliri
Department of Computer Engineering, Karaj Branch, Islamic Azad University,
Karaj, Iran
Arman.Daliri@kiau.ac.ir

***Abstract***

**One of the most important problems in medicine is to facilitate prediction. In this study, we propose entropy triangle method, a novel framework for predicting heart rhythms using a novel machine learning technique. This framework includes three steps: feature engineering, entropy triangle oversampling, and disease prediction. The dataset used in this study is a 12-lead electrocardiogram (ECG) arrhythmia research database with 10,646 patients. This dataset contains 11 different heart rhythms (5 sinus rhythms and 6 non-sinus rhythms). In this article, we introduce two firsts in machine learning and medicine that can predict non-sinus rhythm with over 85% accuracy. Our experimental results show, among others, that the most accurate classifier based on entropy triangles and the most useful oversampling are the supported vector classifiers and oversampling techniques for shark scent.**

## 1. INTRODUCTION

Heart problems are one of the leading causes of death globally. In the United States, up to 11% of people between ages of 20 and 40 years old have cardiovascular disease. Between 40 and 60 years old, the number of patients reaches 37%, and at the age of over 80, it reaches 85% [1]. Heart diseases can be categorized into different types, such as cardiovascular disease [2], coronary artery disease [3], heart failure [4], heart attack [5], and arrhythmia [6]. Sometimes, these problems are caused by a brain or heart attack that may result in death. For these reasons, there has been a great deal of effort and research to predict heart diseases with a high accuracy. There are several methods to evaluate heart performance, and one of the most common ones is the Electrocardiogram records (ECG) [7], which is a test to check the heart arrhythmia [8].

Arrhythmia is one of the most common issues among patients with heart disease. Cardiac arrhythmia is a disorder caused by any changes in the regular heart rhythm, which can be recognized by ECG [9]. Congenital problems, improper lifestyle, and smoking are among the common causes of cardiac arrhythmia. There are several types of arrhythmias, and one of the most important types, which can cause death, can be predicted based on heart atrial rate [10]. Arrhythmias can be categorized into two classes of sinus and non-sinus rhythms [11]. In contrast to sinus rhythms, which are low-risk rhythms, non-sinus rhythms can be considered high risk, which cause death or brain attack [12].

In several articles like [11] the importance of specific problems in the non-sinus class is described. Although specific rhythms play an important role in some heart problems, it is fundamental to investigate all non-sinus rhythms together. In a recent work on importance of left atrial infiltration [13], the researcher's study eleven arrhythmias in detail; however, there is no hint of sinus and non-sinus rhythms. They also did not study the importance of the difference between the sinus and non-sinus rhythms. Our work proposes a method to describe this vital difference and presents a new method to predict the sinus and non-sinus rhythms.

In this research, a framework is presented to predict important types of Cardiac arrhythmia. The research aims to predict sinus and non-sinus arrhythmias using an imbalanced dataset. The novelty of this work is to categorize the heart arrhythmia to sinus and non-sinus classes and to build a new machine learning method to automatically select the best sampler for different types of datasets. This new method is called Entropy Triangle (ET) and its medical novelty is non-sinus rhythms prediction (NRP). We evaluate the performance of the proposed method based on a large variety of predictions and evaluations over various datasets. Among different findings, we report 85% accuracy in prediction of sinus and non-sinus rhythms using our entropy triangle-based framework.

This paper is organized as follows: Section 2 discusses related work and background information about heart disease and types of arrhythmias. Section 3 describes the proposed framework, the Entropy-based method, and its implementation. Section 4 supplies an evaluation of the proposed framework. Finally, Section 5 summarizes the main conclusions.

## 2. Background Information

This section presents an overview on sinus and non-sinus rhythms. First, the types of arrhythmias used in this work are represented are described in detail.

A cardiac arrhythmia occurs when the heart's electrical pulses do not show the heartbeat properly [9]. Three common and well-known arrhythmia models are Tachycardia, Bradycardia, and Fibrillation. In the first Model (Tachycardia), the heart rate is faster than usual; in the second Model (Bradycardia), the heart rate is slower than usual; and in the third Model (Fibrillation), the heart rate beats irregularly [9]. Symptoms of this condition can include difficulty of breathing, palpitations when the heart is beating too fast, lightheadedness, nausea or vomiting, and chest pain. Diagnosis of this disease can be made using an ECG device. While in most cases medicine is used to treat the problem, as the disease worsens, radio frequencies and, in severe conditions and when the above methods are not effective, heart surgery are accomplished.

Cardiac arrhythmia is recorded using an electrodiagram device, and physicians can notice arrhythmic abnormalities relevant to the recorded signals. An electro diagram (ECG) device prints signals with different nodes, including heart rate [14]. Heart rhythm contains a graph called PQRST. Figure 1 indicates a heart rhythm, which begins from P. A small segment is represented by PR after the P signal. Then, the heart rhythm reaches the heart complex, shown by QRS Complex. Finally, an ST segment begins, which ends with a T rhythm [14].

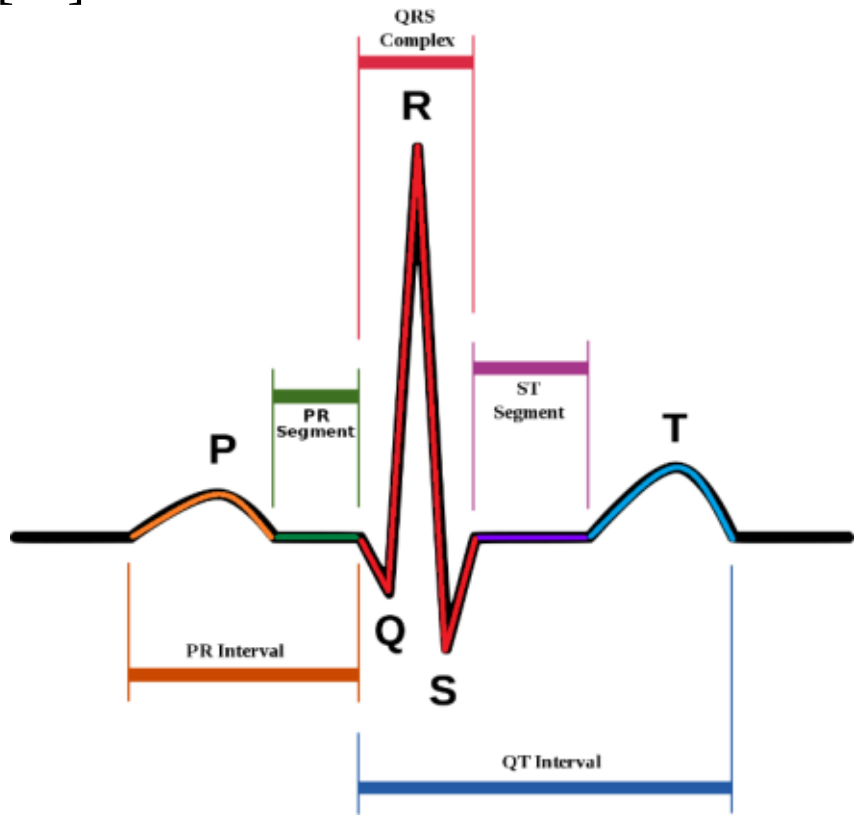

*Figure 1. PQRST Heart rhythm*

sinus arrhythmias are disorders that are not life-threatening. In this disorder, the heart rhythm may be slower or faster than usual, but it does not pose a severe threat to the patient's life [15]. The sinus rhythm (SR) is the most common heart rhythm in people who do not have an arrhythmia [16]. Sinus bradycardia (SB) is another sinus rhythm in which the heart beats more slowly than usual, meaning that the standard heart rate beats between 60 and 100 beats per minute, but in SB, the heart beats below 60 beats [17]. Sinus tachycardia (ST) is another arrhythmic disease in which the heart beats continuously and more than usual in one minute [18]. In Sinus Irregularity (SI), heart is sinusoidal but irregularly spaced 10 to P apart from the P-to-P wave distance [19]. Another arrhythmia used in this article is Sinus Atrium to Atrial Wandering Rhythm (SAAWR). In this type of disease, heartbeat is disturbed in sinus rhythms in the P wave part [19]. This disease is observed in very young and significantly older people.

The non-sinusoidal and high-risk rhythms studied in this article include six different rhythms. Atrial Fibrillation (AFIB) is one of the most common non-sinus rhythms. This disease is hazardous, and the heart rhythm is irregular. AFIB disease can lead to blood clots in brain or heart failure [20]. The second non-sinusoidal rhythm is Atrial Flutter (AF), which is hazardous and occurs when a rapid and irregular heartbeat appears in the heart's chambers. This disease can lead to stroke, which has a lasting effect on patients' lives [21]. Another dangerous disease is Atrial Tachycardia (AT), which produces an abnormal rhythm. Such rhythms show an abnormal beat from the atria and recur rapidly [22]. Another type of non-sinusoidal rhythm is Atrioventricular Node Reentrant Tachycardia (AVNRT), a type of Supraventricular Tachycardia (SVT). SVT is a low-risk non-sinus rhythm that can be controlled with drug therapy. The most common Model of this type of rhythm is AVNRT [23]. Atrioventricular Reentrant Tachycardia (AVRT) is slightly more dangerous than SVT and has a heart rate of more than 100 beats per minute [24]. All the said heart rhythms have been examined, tested, and evaluated in this study.

## 3. Methodology of The Entropy Triangle Method (ETM)

In this section, our novel machine learning method, Entropy Triangle (ET), and the structure of the proposed framework are explained. First of all in part A, an overview of the framework, and then in part B, the Entropy Triangle (ET) method describes. Finally in part C, medical novelty of the proposed method, which is Non-Sinus and Sinus Rhythm Prediction, is further discussed.

### *A. Overview of the framework*

The proposed framework aims to determine heart sinus and non-sinus rhythms. To do so, it utilizes supervised machine learning methods, and uses three steps to distinguish sinus and non-sinus rhythms. The first step, feature engineering, includes three parts: data collection, data de-identification, and feature selection. Step 2, Entropy Triangle-Based Oversampling (ETBO), shows a novel algorithm we propose in this work. In ETBO, three metrics are employed to select a proper method of oversampling. Finally, step 3, Non-Sinus and Sinus Rhythm Prediction, predicts non-sinus and sinus rhythms.
As shown in Figure 2, the first step is comprised of three parts: data collection, data de-identification, and feature selection. The focus of data collection is to gather valuable data and features of heart rhythm legally and ethically. This choice is based on the advice of a heart specialist and the database used in [14]. Next, based on the data information, personally identifying information of every 10,000 patients was removed to protect their privacy. Finally, according to previous studies on heart rhythm [14] valuable features are selected for the next steps. As the topic of this research is the prediction of the heart rhythm, rhythm is considered the target feature in our analyses.

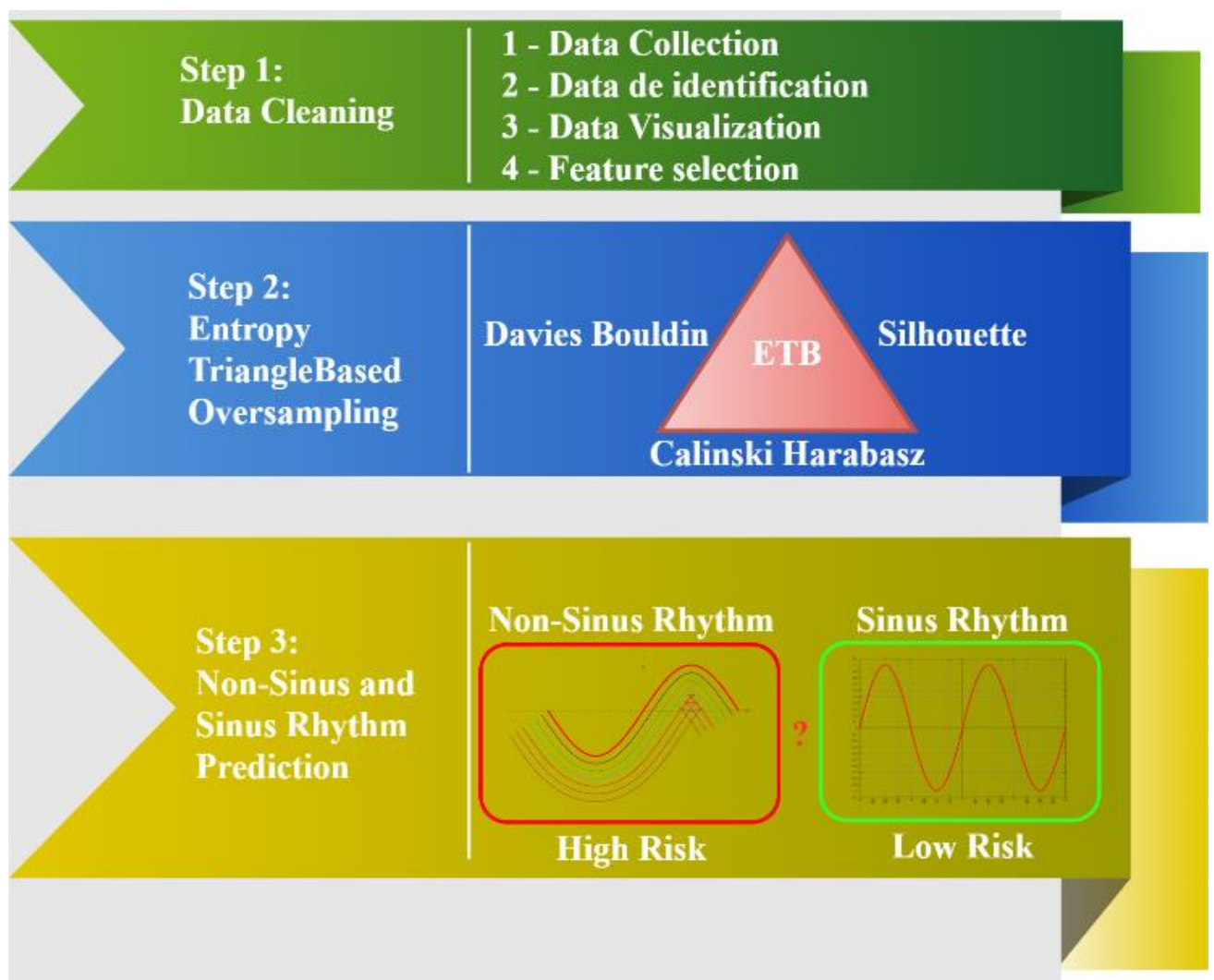

*Figure 2. Steps of Entropy triangle-based framework*

### B. Entropy Triangle Based Oversampling

Entropy Triangle Based Oversampling (ETBO) focuses on choosing the best oversampler for balancing the dataset. The issue with an imbalanced data is revealed when we train and test the classifier algorithms. As one of the classes is minority, the classifier cannot gain a large enough or a representative sample of the minority class [25]. Several oversampling methods have been proposed to solve this issue [26], [27], and ETBO intends to select the best one for the data used in this work.
ETBO works based on three numerical metrics to select the best oversampling method from a set of oversampling candidates. For each candidate, the values of these three metrics are computed and used to make three sides of a triangle. Then, the area of each triangle is calculated and compared with the areas of other candidates. The three metrics used in ETBO are adjusted in a way that the best oversampling method will be the one with the lowest triangle area. The three metrics used in ETBO are Calinski-Harabasz score [28], silhouette score [29], and Davies-Bouldin score [30], which are described in detail as below:
Calinski-Harabasz uses the K-means algorithm to predict the suitable classes in a dataset [28]. Calinski-Harabasz index is the ratio of clusters dispersion over the sum of within-cluster dispersion for each cluster. Value of this metric is higher when clusters are dense. Silhouette score uses the K-means algorithm to define the suitable classes of a dataset [29]. This metric is based on two scores to evaluate each sample. The value of this metric is between -1 to +1, and the maximum value means that all the classes are clearly distinguished. Davies-Bouldin score, same as the previous metrics, uses the K-means algorithm [30]. This metric's best value is zero, which means that classes are ideally identified. The main focus of the Davies-Bouldin score is on the average similarity between classes.
To create the ETBO triangle, the three metrics are computed for each oversampling method, and a triangle is created using the three numerical values. If an oversampling candidate could distinguish the clusters clearly, the values of all the three metrics create a triangle with the lowest area. The best value of Silhouette score is when it's between -1 and 1. Calinski-Harabasz score value is upper than 1 and is a ratio of within-cluster and between-cluster dispersions.

### C. Rhythm Prediction

After running ETBO, classification algorithms are applied, and the prediction is made. In Section 4, we conduct a comprehensive experiment over the proposed framework and evaluate its performance in predicting rhythms. As mentioned before, heart regular rhythm is sinusoidal, and when there are some problems in body, the rhythm will become non-sinusoidal. Problems with the pacemaker nodes in the heart [31], and atrial or ventricular problems [32] can cause arrhythmia. Arrhythmia can also due to blockage of the heart arteries [33], or myocardial infarction and ischemia [34] that can cause heart attack or brain attack. The non-sinusoidal heart rhythms can threaten patients' life. One of the most dangerous non-sinus rhythms is the atrial fibrillation rate. The experiments presented in Section 4 aim to lower the risk of such issues by predicting sinus and non-sinus rhythms, which is the novelty of this work in the field of medical science.
It is worth mentioning that ECG is not always a valid resource of finding sinus or non-sinus rhythms. For instance, some medicines can cause non-sinus rhythms. In such cases, physicians check for the medication, the range of blood electrolytes, and the range of potassium that can have immediate effects on rhythms [35]. One of the most common cases is when a patient is dehydrated for severe vomiting that affects the potassium range [36]. Also, sympathetic problems can cause arrhythmia in a patient. Such problems can affect sinus rhythms, and when an ECG is recorded, it may be non-sinus rhythms.

## 4. Experimental Results

This section reports our extensive experiments to evaluate the proposed framework. First, the dataset used in this work is described. Next, a detailed information on selecting the best oversampling method using the proposed entropy triangle (ET) method is presented. Then, we provide the evaluation results of sinus and non-sinus rhythm prediction.

### Data set

To evaluate our proposed algorithm, we use a 12-lead electrocardiogram (ECG) database for arrhythmia research that contains the data of 10,646 patients [14]. These data are collected during a 10-second resting ECG test at Chapman University and Shaoxing People's Hospital. The database reports the heart-related information of patients who suffered from eleven types of arrhythmias.

The data employed in this article uses the rhythm feature to categorize different rhythms. To have clearer classes, based on a cardiologist's recommendation, 11 different heart rhythms in this data have been divided into two categories, and then the classification has been applied. Table 1 indicates more detail of sinus and non-sinus rhythms. The five existing sinus rhythms of Sinus Rhythm (SR), Sinus Bradycardia (SB), Sinus Tachycardia (ST), Sinus Irregularity (SI), and Sinus Atrium to Atrial Wandering Rhythm (SAAWR) are placed in one category and are introduced as sinus rhythms in the classification. The second category, non-sinusoidal arrhythmia, includes six rhythms named Atrial Fibrillation (AFIB), Atrial Flutter (AF), Supraventricular Tachycardia (SVT), Atrial Flutter (AT), Atrioventricular Node Reentrant Tachycardia (AVNRT), and Atrioventricular Reentrant Tachycardia (AVRT). The main factor for separating 11 different rhythms is the elevated risk of most non-sinusoidal rhythms. Although sinus rhythms are also dangerous, they pose less risk to patients' life than non-sinus rhythm

*Table 1.Collection of targets*

| New collection | Merged from | Count | Testing data size (20%) | Training data size (80%) |
|---|---|---|---|---|
| Sinus Rythm | SR, SB, ST, SI, SAAWR | 7735 | 2340 | 6176 |
| Non-sinus Rythm | AFIB, AF, SVT, AT, AVNRT, AVRT | 2911 | 571 | 1559 |

***Balancer***

After preparing the categorical targets, we use balancer methods to balance the data. Localized Random Affine Shadow sampling (LORAS) [37], synthetic minority over-sampling technique (SMOTE) [38], random oversampling (ROS) [39], synthetic minority over-sampling technique with Extended nearest neighbor (SMOTE-ENN) [40], SMOTE TOMEK (ST) [41], Cluster SMOTE (CS) [42], a combined cleaning and resampling algorithm for imbalanced data classification (CCR) [43], and Shark smell optimization (SSO) [44] are the balancing methods used in our experiments.

Table 2 lists the triangle area for each balancer method after applying the ETBO algorithm, and Figure 3 plots the area entropy triangles. Based on the results, SSO has the lowest triangle area equal to 0.4065, and CS with a notable difference is ranked second. ROS with the area of 0.7001 has the maximum area, and the rest, CCR, SMOTE-ENN, and ST, have areas close to the maximum. For the further experiments, we employ SSO as the balancer since it has the minimum entropy triangle area.

*Table 2.Entropy triangle result of each balancer*

| Balancer | LORAS | SMOTE | ROS | SMOTE-ENN | ST | CS | CCR | SSO |
|---|---|---|---|---|---|---|---|---|
| **Triangle Area** | 0.6297 | 0.6754 | 0.7001 | 0.6938 | 0.6802 | 0.5277 | 0.6973 | **0.4065** |

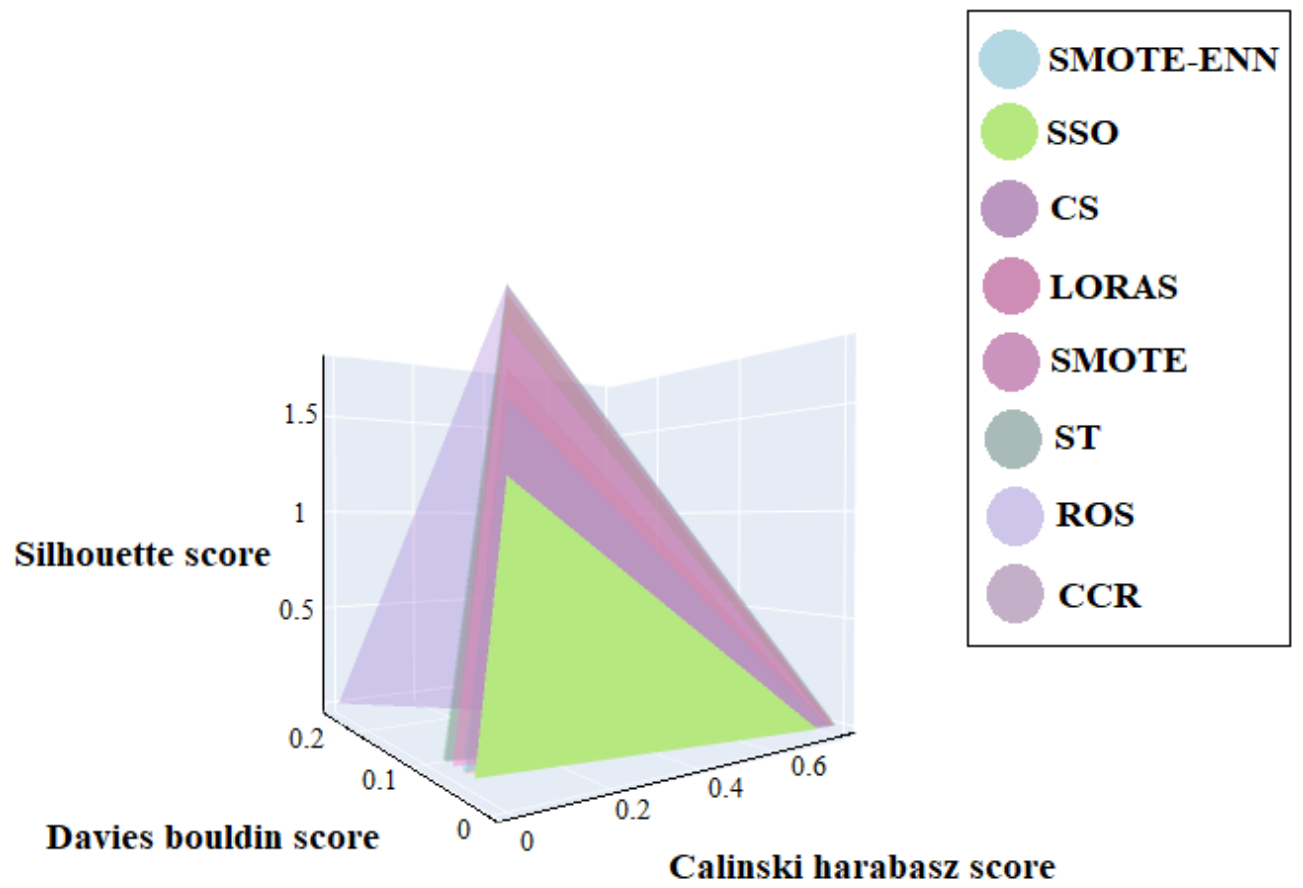

*Figure 3.Area entropy triangles in a plot with origin of (0, 0, 0)*

***Classification***

Now, we evaluate the performance of three classifiers using the data balanced in section 4.2. The goal is to explore an accurate classification of sinus and non-sinus rhythms. Based on subsection 4.2, the best balancer method is SSO, and the classification algorithms are evaluated on the data balanced by this method. In the evaluation, the data are divided into 80% for training and 20% for testing. The classifiers we use are, Logistic Regression (LR) [45], Support Vector Machine (SVC) [46], and Naive bayes (NB) [47]. All initial parameters for each classifier are set to the default values. the metrics we used to evaluate the performance of the classifiers. The raw data used in the experiments include 10646 patients, 7735 of which belong to the sinus rhythm group, and the rest are in the non-sinus rhythm group. Table 3 indicates the evaluation metrics, F1-score, precision, and recall, for eight different balancers and three classification methods. The first row presents the performance of the classifiers over the original data that is imbalanced. The results indicate a better performance for the majority class, which is the sinus rhythm. However, the focus of this work is to provide an accurate classification for both sinus (the majority class) and non-sinus (the minority class) rhythms.

The rest of the table indicates the classification performance when the data is balanced by the eight balancer methods. All balancers affect the evaluation results, especially the minority class, the non-sinus rhythm group. As the results show, classification over the data balanced by SSO results in a better performance in contrast to the others. SVC and NB have the highest F1-score in comparisons with the other classification methods. For the other evaluation metrics, the results of these two classifiers are approximately close to each other. It is worth mentioning that some classifiers have a higher evaluation metric than SVC and NB. However, the overall performance of these two methods for both class of sinus and non-sinus rhythms is higher than the others.

*Table 3.F1-score, precision (Pr) and recall (R) for three classifiers over imbalanced and balanced data.*

| Balancer | Classifier | F1 of sinus rhythm | F1 of non-sinus rhythm | Pr of sinus rhythm | Pr of non-sinus rhythm | R of sinus rhythm | R of non-sinus rhythm |
|---|---|---|---|---|---|---|---|
| Raw data | LR | 86% | 52% | 82% | 63% | 90% | 44% |
| | SVC | 88% | 53% | 82% | 75% | 95% | 42% |
| | NB | 85% | 58% | 84% | 59% | 86% | 56% |
| **SSO** | LR | 80% | 80% | 80% | 80% | 80% | 80% |
| | **SVC** | **86%** | **85%** | **82%** | **89%** | **89%** | **81%** |
| | **NB** | **86%** | **84%** | **81%** | **90%** | **89%** | **81%** |
| LORAS | LR | 73% | 72% | 72% | 73% | 74% | 70% |
| | SVC | 77% | 76% | 77% | 76% | 77% | 76% |
| | NB | 73% | 66% | 67% | 75% | 81% | 59% |
| SMOTE | LR | 73% | 73% | 73% | 73% | 74% | 72% |
| | SVC | 77% | 76% | 77% | 76% | 77% | 76% |
| | NB | 74% | 67% | 67% | 76% | 82% | 59% |
| ROS | LR | 73% | 72% | 73% | 80% | 83% | 69% |
| | SVC | 77% | 76% | 76% | 77% | 78% | 75% |
| | NB | 74% | 67% | 67% | 76% | 82% | 59% |
| S-ENN | LR | 78% | 80% | 79% | 80% | 77% | 81% |
| | SVC | 80% | 82% | 80% | 81% | 79% | 82% |
| | NB | 75% | 73% | 69% | 80% | 81% | 68% |
| S-T | LR | 74% | 73% | 75% | 73% | 73% | 74% |
| | SVC | 74% | 75% | 76% | 73% | 73% | 76% |
| | NB | 72% | 66% | 67% | 73% | 78% | 60% |
| C-S | LR | 77% | 77% | 77% | 77% | 78% | 76% |
| | SVC | 81% | 80% | 79% | 82% | 84% | 77% |
| | NB | 79% | 75% | 73% | 82% | 85% | 69% |
| CCR | LR | 72% | 74% | 72% | 74% | 72% | 74% |
| | SVC | 74% | 76% | 75% | 75% | 72% | 78% |
| | NB | 73% | 69% | 66% | 78% | 81% | 61% |

## 5. CONCLUSION

This article describes the early prevention of heart arrhythmia based on sinus and non-sinus rhythms. We propose a novel framework that uses an entropy triangle to select the best balancer for the data. We provide several experiments to evaluate the framework using both balanced and imbalanced data. The results indicate that our framework predicts non-sinus rhythms with an accuracy of 85%. The novelties of this work could provide inspiration for researchers in both machine learning and medical studies.

For future work, we plan to evaluate our framework using different big data and study its performance for other types of disease in the healthcare field. Improving the set of balancers and classifiers in the framework could be also another direction for the future work. An online version of the proposed framework could be implemented and used for runtime cardiac arrhythmia prediction.